\pdfoutput=1

\documentclass[11pt]{article}

\usepackage{acl}

\usepackage{times}
\usepackage{latexsym}

\usepackage[T1]{fontenc}

\usepackage{amsmath,amsfonts,bm}

\def\eqref#1{equation~\ref{#1}}

\def\1{\bm{1}}

\DeclareMathAlphabet{\mathsfit}{\encodingdefault}{\sfdefault}{m}{sl}
\SetMathAlphabet{\mathsfit}{bold}{\encodingdefault}{\sfdefault}{bx}{n}

\newcommand{\E}{\mathbb{E}}

\usepackage[utf8]{inputenc}
\usepackage[T1]{fontenc}
\usepackage{rotating,tabularx}
\usepackage{multirow}
\usepackage{float}
\usepackage{booktabs} 
\usepackage{url}
\usepackage{graphicx}  
\usepackage{amsfonts} 
\usepackage{amsmath}
\usepackage{bm}
\usepackage{multirow}
\usepackage{subfigure}
\usepackage{algorithm}  
\usepackage{algpseudocode}  
\usepackage{makecell}
\usepackage{xspace}
\usepackage{amssymb}
\usepackage{array}
\usepackage{footmisc}
\usepackage{listings}
\usepackage[np, autolanguage]{numprint}
\usepackage{paralist}
\usepackage{enumitem}
\usepackage{colortbl}
\usepackage{soul}
\usepackage{epstopdf}
\usepackage{pifont}
\usepackage{xcolor}
\usepackage{mathrsfs}
\usepackage{todonotes}
\usepackage{graphicx}
\usepackage{CJK}
\newcommand{\dubbelop}{$^{\blacktriangle}$}
\newcommand{\hlc}[2][yellow]{{%
    \colorlet{foo}{#1}%
    \sethlcolor{foo}\hl{#2}}%
}
\usepackage{microtype}
\newcommand{\ie}{\emph{i.e.,}\xspace}

\title{Keywords and Instances: A Hierarchical Contrastive Learning Framework Unifying Hybrid Granularities for Text Generation}

\author{Mingzhe Li\textsuperscript{\rm 1,2,3,}\thanks{\ \ \ Equal Contribution}\ , Xiexiong Lin\textsuperscript{\rm 3,}\footnotemark[1]\ , Xiuying Chen\textsuperscript{\rm 4}, Jinxiong Chang\textsuperscript{\rm 3}, Qishen Zhang\textsuperscript{\rm 3},\\ {\bf Feng Wang\textsuperscript{\rm 3}}, {\bf Taifeng Wang\textsuperscript{\rm 3}}, {\bf Zhongyi Liu\textsuperscript{\rm 3}}, {\bf Wei Chu\textsuperscript{\rm 3}},
{\bf Dongyan Zhao\textsuperscript{\rm 1,2,}\thanks{\ \ \ Corresponding authors: Rui Yan and Dongyan Zhao}}\ ,
{\bf Rui Yan\textsuperscript{\rm 5,}\footnotemark[2]}\\
$^1$Wangxuan Institute of Computer Technology, Peking University, Beijing, China\\
$^2$Center for Data Science, AAIS, Peking University, Beijing, China\\
$^3$ Ant Group \  $^4$ Computational Bioscience Reseach Center, KAUST\\
$^5$ Gaoling School of Artificial Intelligence, Renmin University of China\\
 \texttt {li\_mingzhe@pku.edu.cn},\  \texttt {xiexiong.lxx@antfin.com}
}

\begin{document}
\maketitle
\begin{abstract}
Contrastive learning has achieved impressive success in generation tasks to militate the ``exposure bias'' problem and discriminatively exploit the different quality of references.
Existing works mostly focus on contrastive learning on the instance-level without discriminating the contribution of each word, while keywords are the gist of the text and dominant the constrained mapping relationships.
Hence, in this work, we propose a hierarchical contrastive learning mechanism, which can unify hybrid granularities semantic meaning in the input text.
Concretely, we first propose a keyword graph via contrastive correlations of positive-negative pairs to iteratively polish the keyword representations.
Then, we construct intra-contrasts within instance-level and keyword-level, where we assume words are sampled nodes from a sentence distribution.
Finally, to bridge the gap between independent contrast levels and tackle the common contrast vanishing problem, we propose an inter-contrast mechanism that measures the discrepancy between contrastive keyword nodes respectively to the instance distribution. 
Experiments demonstrate that our model outperforms competitive baselines on paraphrasing, dialogue generation, and storytelling tasks\footnote{Our model has been integrated into the search feature on the homepage of the Alipay app.}.
\end{abstract}

\section{Introduction}
\label{introduction}

Generation tasks such as storytelling, paraphrasing, and dialogue generation aim at learning a certain correlation between text pairs that maps an arbitrary-length input to another arbitrary-length output.
Traditional methods are mostly trained with ``teacher forcing'' and lead to an ``exposure bias'' problem~\citep{schmidt2019generalization}.
Incorporating the generation method with contrastive learning achieved impressive performance on tackling such issues, which takes an extra consideration of synthetic negative samples contrastively \citep{Claps}.

\begin{figure}[t]
\centering
\includegraphics[scale=0.62]{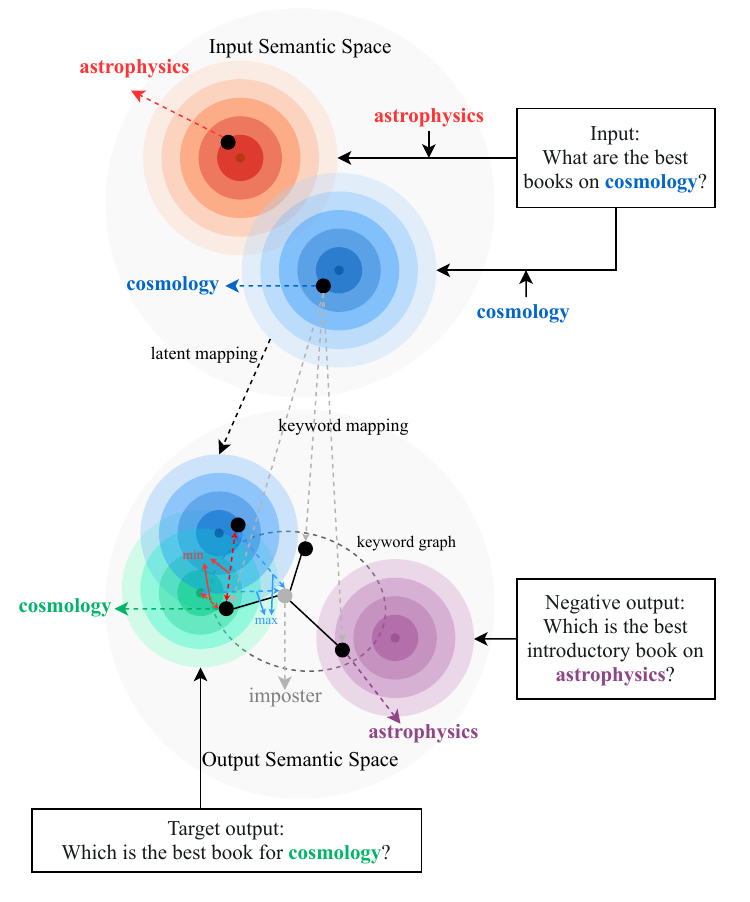}
\caption{
    The semantic meaning of the sentence ``what are the best books on cosmology?'' would be greatly changed if the keyword ``cosmology'' is changed to ``astrophysic''.}
\label{fig:intro}
\end{figure}

Existing contrastive mechanisms are mainly focused on the instance level~\citep{Claps,group-wise}.
However, word-level information is also of great importance.
Take the case shown in the upper part of Figure \ref{fig:intro} for example, the keyword covers the gist of the input text and determines the embedding space of the text.
The text representation will be significantly affected if adding a slight perturbation on the keyword, \ie changing ``cosmology'' to ``astrophysics''.
In addition, as shown on the bottom part, under some circumstances, it is too easy for the model to do the classification since the semantic gap between contrastive pairs is huge.
Thus, the model fails to distinguish the actual discrepancy, which causes a ``contrast vanishing'' problem at both instance-level and keyword-level.

Based on the above motivation, in this paper, we propose a hierarchical contrastive learning method built on top of the classic CVAE structure.
We choose CVAE due to its ability in modeling global properties such as syntactic, semantic, and discourse coherence \citep{li2015hierarchical, yu2020draft}.
We first learn different granularity representations through two independent contrast, \ie instance-level and keyword-level.
Specifically, we use the universal and classic TextRank~\cite{mihalcea2004textrank} method to extract keywords from each text, which contain the most important information and need to be highlighted.
On the instance-level, we treat the keyword in the input text as an additional condition for a better prior semantic distribution.
Then, we utilize Kullback–Leibler divergence~\cite{kullback1951information} to reduce the distance between prior distribution and positive posterior distribution, and increase the distance with the negative posterior distribution. 
While on the keyword-level, we propose a keyword graph via contrastive correlations of positive-negative pairs to learn informative and accurate keyword representations.
By treating the keyword in the output text as an anchor, the imposter keyword is produced by neighboring nodes of the anchor keyword and forms the keyword-level contrast, where the similarity between the imposter keyword and the anchor keyword is poorer than the positive keyword.

To unify individual intra-contrasts and tackle the ``contrast vanishing'' problem in independent contrastive granularities, we leverage an inter-contrast, the Mahalanobis contrast, to investigate the contrastive enhancement based on the Mahalanobis distance~\citep{de2000mahalanobis}, a measure of the distance between a point and a distribution, between the instance distribution and the keyword representation.
Concretely, we ensure the distance from the anchor instance distribution to the ground-truth keyword vector is closer than to the imposter keyword vector.
The Mahalanobis contrast plays an intermediate role that joins the different granularities contrast via incorporating the distribution of instance with the representation of its crucial part, and makes up a more comprehensive keyword-driven hierarchical contrastive mechanism, so as to ameliorate the generated results.

We empirically show that our model outperforms CVAE and other baselines significantly on three generation tasks: paraphrasing, dialogue generation, and storytelling. 

Our contributions can be summarized as follows:

$\bullet$ \textcolor{black}{To our best knowledge, we are the first to propose an inter-level contrastive learning method, which unifies instance-level and keyword-level contrasts in the CVAE framework.}

$\bullet$ \textcolor{black}{We propose three contrastive learning measurements: KL divergence for semantic distribution, cosine distance for points, and Mahalanobis distance for points with distribution.}

$\bullet$ \textcolor{black}{We introduce a global keyword graph to obtain polished keyword representations and construct imposter keywords for contrastive learning.}

\section{Related Work}
\label{rela_work}

\subsection{Contrastive Learning}
Contrastive learning is used to learn representations by teaching the model which data points are similar or not.
Due to the excellent performance on self-supervised and semi-supervised learning, it has been widely used in natural language processing (NLP).
Firstly, \citet{mikolov2013distributed} proposed to predict neighboring words from context with noise-contrastive estimation.
Then, based on word representations, contrastive learning for sentence has been utilized to learn semantic representations.
\citet{Claps} generated positive and negative examples by adding perturbations to the hidden states.
\citet{group-wise} augmented contrastive dialogue learning with group-wise dual sampling.
Moreover, contrastive learning has also been utilized in caption generation~\citep{mao2016generation}, summarization~\citep{liu2021simcls} and machine translation~\citep{yang2019reducing}.
Our work differs from previous works in focusing on hierarchical contrastive learning on hybrid granularities.

\subsection{Mahalanobis Distance}
The Mahalanobis distance is a measure of the distance between a point and a distribution~\citep{de2000mahalanobis}.
The distance is zero if the point is on the distribution.
Recently, Mahalanobis distance is popularly applied to the NLP tasks \citep{tran2019adversarial}.
\citet{podolskiy2021revisiting} showed that while Transformer is capable of constructing homogeneous representations of in-domain utterances, the Mahalanobis distance captures geometrical disparity from out of domain utterances.
Further, \citet{ren2021simple} considered that the raw density from deep generative models may fail at out-of-domain detection and proposed to fix this using a likelihood ratio between two generative models as a confidence score.

\subsection{Conditional Variational Auto-Encoder}
Variational autoencoder (VAE) was proposed by~\citet{kingma2013auto}, and has been widely used in various tasks such as headline generation~\citep{li2021style}, dialogue generation~\citep{serban2017hierarchical} and story generation~\citep{yu2020draft}.
Based on VAE, a more advanced model, Conditional VAE (CVAE), was proposed to generate diverse images conditioned on certain attributes, which was also applied to generate diverse outputs in NLP tasks \citep{zhao2017learning, qiu2019training}.
Existing works concentrate on generating diverse outputs, and we take one step further to utilize prior and posterior latent distribution to compare positive and negative samples, which helps to learn more accurate semantic information.
\section{Method}
\label{method}

\begin{figure*}
\centering
\includegraphics[scale=0.68]{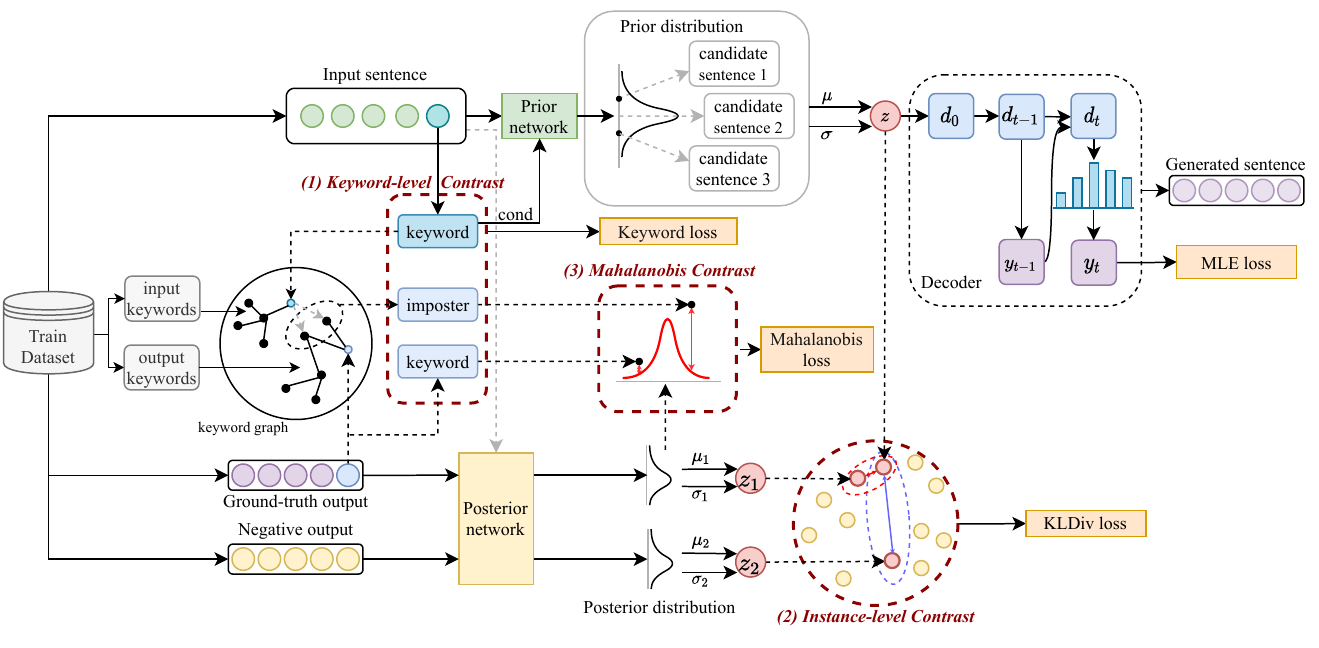}
\caption{
    The architecture of hierarchical contrastive learning, which consists of three parts: (1) Keyword-level contrast from keyword graph; (2) Instance-level contrast based on KL divergence for semantic distribution; and (3) Mahalanobis contrast between instance-level and keyword-level.
}
\label{fig:model}
\end{figure*}

\subsection{Background}
\paragraph{VAE:} Variational auto-encoder (VAE) is a typical encoder-decoder structural model with certain types of latent variables. 
Given an input $x$,
VAE models the latent variable $z$ through the prior distribution $p_\theta(z)$ , and the observed data $x$ is reconstructed by the generative distribution $p_\theta(x|z)$ which is the likelihood function that generates $x$ conditioned on $z$. 
Since $z$ is unknown, it should be estimated according to the given data $x$ as $p_\theta(z|x)$.
While the posterior density $p_\theta(z|x)=p_\theta(x|z)p_\theta(z)/p_\theta(x)$ is intractable, VAE introduces a recognition posterior distribution $q_\phi(z|x)$ approximates to the true posterior $p_\theta(z|x)$. 
Thus, VAE is trained by optimizing the lower bound on the marginal likelihood of data $x$ as:
\begin{equation}
\label{vae}
\begin{split}
    log p_\theta(x) \geq \E_{z\sim q_\phi(z|x)}[log p_\theta(x|z)] \\ - D_{KL}(q_\phi(z|x) || p_\theta(z)),
\end{split}
\end{equation}
where $D_{KL}$ is the Kullback–Leibler divergence. 

\paragraph{CVAE:} The conditional variational auto-encoder (CVAE) is the supervised version of VAE with an additional output variable.
Giving a dataset $\{x_i, y_i\}_{i=1}^N$ consisting of $N$ samples,
CVAE is trained to maximize the conditional log-likelihood, and the variational lower bound of the model is written as follows:
\begin{equation}
\begin{split}
    log p_\theta(y|x) \geq \E_{z\sim q_\phi(z|x,y)}[log p(y|x,z)] \\
    - D_{KL}(q_\phi(z|x,y) || p_\theta(z|x)).
\end{split}
\end{equation}
Assuming the type of latent variable obeys Gaussian distribution, the first right-hand side term can be approximated by drawing samples $\{z_i\}_{i=1}^N$ from the recognition posterior distribution $q_\phi(z|x,y)$, where $z\sim N(\mu, \sigma^2I)$, and then objective of the CVAE with Gaussian distribution can be written as:
\begin{equation}
\label{cvae_rp}
    \begin{split}
    \mathcal{L}_{cvae}(x,y;\theta,\phi) =  -\frac{1}{N} \sum^{N}_{i=1} log p_\theta (y|x,z_i)
    \\  + D_{KL}(q_\phi(z|x,y) || p_\theta(z|x)),
    \end{split}
\end{equation}
where $z_i=g_\phi(x, y, \epsilon_i)$, $\epsilon_i \sim \mathcal{N}(0, I)$. 
The distribution $q_\phi(z|x,y)$ is reparameterized with a differentiable function $g_\phi$, which enables the model trainable via stochastic gradient descent.

Inspired by \citet{wu2019guiding}, we add keyword $u$ as an additional condition to the prior distribution to control the generation process, which turns the $p_\theta(z|x)$ in Equaton~\ref{cvae_rp} into $p_\theta(z|x,u)$.

\subsection{Hierarchical Contrastive Learning}
In this section, we introduce our hierarchical contrastive learning method, which is comprised of three parts: instance-level contrast based on KL divergence (sec.\ref{cl_ins}),  keyword-level contrast based on keyword graph (sec.\ref{keyword_cont}), and inter-contrast: Mahalanobis contrast (sec.\ref{ma_dis}).

\subsubsection{Instance-level Contrastive Learning}
\label{cl_ins}
To tackle the ``exposure bias'' problem and discriminatively exploit the different quality of references, instance-level contrastive learning is introduced to learn discrepancies of targets.
Specifically, in addition to the observed input data $x$ and positive output $y^+$, a negative output $y^-$ is added to construct a contrastive pair $\{(x,y^+),(x,y^-)\}$.
In this case, the prior distribution $p_\theta(z|x)$ is learned from a prior network, which is denoted as $f_\theta(x)$.
The approximate posteriors $q_\phi(z|x,y^+)$ and $q_\phi(z|x,y^-)$ are learned from a posterior network and represented as $f_\phi(x,y^+)$ and $f_\phi(x,y^-)$, respectively.
The objective here is to make the distance between a prior distribution and positive posterior distribution closer than with the negative posterior distribution.
Thus, the instance-level contrastive loss function can be written as:
\begin{align*}
\setlength{\abovedisplayskip}{1pt}
\small
\begin{split}
    \mathcal{L}_{ins} =  
    -\E_{f_\phi}[log(1-\frac{e^{h(f_\phi(x,y^+), f_\theta(x)) / \tau}}{\sum_{y^* \in Y}e^{h(f_\phi(x,y^*), f_\theta(x))/\tau}})],
\end{split}
\end{align*}
where the $y^* \in Y$ can be positive sample $y^+$ or negative sample $y^-$, and the $\tau$ is a temperature parameter to control push and pull force.
The function $h(\cdot)$ denotes the distance between elements, which is set as Kullback–Leibler divergence~\citep{kullback1951information} in instance-level contrast, $D_{KL}(f_\phi(x,y^*)||f_\theta(x))$, to measure the difference between two distributions.

\subsubsection{Keyword-level Contrastive Learning}
\label{keyword_cont}
Since the instance-level contrast focuses on learning high-level information and fails to discriminate the contribution of each word, we incorporate it with a keyword-level contrast to pay more attention to the specific keyword.

\paragraph{Keyword Graph:}
Given an input-output text pair $(x,y)$, keywords $k_x, k_y$ can be extracted from $x$ and $y$, respectively.
For an input text $x_i$ with keyword $k_{x,i}$, input texts that contain the same keyword are gathered into a cluster $C_i=\{x_j\}_{j=1}^n, k_{x,j} \in x_j$, where $n$ is the number of texts in $C_i$.
Each text $x_j\in C_i$ has a positive-negative output text pair $\{(y_j^+, y_j^-)\}$ containing a positive output keyword $k_{y,j}^+$  and a negative one $k_{y,j}^-$, respectively.
Thus, spreading to the entire cluster $C_i$, for the output text $y_i$, there exists positive relations $r^+_{i,j}$ between its keyword $k_{y,i}$ and each of the surrounded positive keywords $\{k_{y,j}^+\}_{j=1}^n$.
Likewise, negative relations $r^-_{i,j}$ correlates the output keyword $k_{y,i}$ and the surrounded negative ones $\{k_{y,j}^-\}_{j=1}^n$.

Based on these keywords as nodes and their relations as edges~\citep{chen2021capturing}, the keyword graph $\mathcal{G}_k$ is constructed.
Each node representation $h_i^{0}$ is initialized as the average BERT embedding~\citep{bert} of texts in the cluster $C_i$ with the same corresponding keyword $k_{x,i}$.
Then, the relation edge $r^{0}_{ij}$ that connects node $i$ and node $j$ is learned via a feedforward layer $r^0_{ij}=\text{FFN}([h^0_i; h^0_j])$.

Then, the representations of nodes and relation edges are iteratively updated with their connected nodes via the graph attention (GAT) layer and the feed-forward (FFN) layer.
In the $t$-th iteration, we first update each edge representation by paying attention to the connected nodes, denoted as:
\begin{gather}
    \beta_{r*}^t = \text{softmax}(\frac{(r_{ij}^t W_p)(h_{*}^{t} W_h)^T}{\sqrt{d}}),\\
    p_{ij}^{t} = \beta_{ri}^t h_{i}^{t} + \beta_{rj}^t h_{j}^{t},\\
    r_{ij}^{t+1} = \text{FFN}(r_{ij}^{t}+p_{ij}^{t}),
\end{gather}
where $h_{*}^{t}$ can be $h_{i}^{t}$ or $h_{j}^{t}$.

Then, based on the obtained edge representation $r_{ij}^{t+1}$, we update the node representations considering both the related nodes and relation edges by the graph attention layer, $\text{GAT}(h_i^t, h_j^t, r_{ij}^t)$, which is designed as:
\begin{align}
    e_{ij}^{t}& = \textstyle \frac{(h_i^{t} W_q)(h_j^{t} W_k + r_{ij}^{t+1} W_r)^T}{\sqrt{d}},\\
    \alpha_{ij}^t &= \textstyle \frac{exp(e_{ij}^t)}{\sum_{l \in N_i}{exp(e_{il}^t)}},\\
    u_i^t &= \textstyle \sum_{j \in N_i}\alpha_{ij}^t ( h_j^t W_v+ r_{ij}^{t+1}),
\end{align}
where $W_q, W_k, W_r$ and $W_v$ are all learnable parameters, and the $\alpha_{ij}^t$ is the attention weight between $h_i^t$ and $h_j^t$.
Besides, to avoid gradient vanishing after several iterations, a residual connection is added to the output $u_i^t$ and the updated node representations $h_i^{t+1}$ is obtained.
In this way, the new representation of each keyword node consists of the relation dependency information from neighbor nodes $N_i$.
We take the node representations from the last iteration as the final keyword representations, denoted as $u$ for brevity.

\paragraph{Keyword-level Contrast:}
The keyword-level contrastive learning arises from input keywords against positive output keywords and negative impostor keywords. 
The input keyword $u_{in}$ is extracted from the input text as an anchor, and the output keyword $u_{out}$ is extracted from ground-truth output text.
While the impostor keyword is calculated from the negative neighbours of the output keyword $u_{out}$,  written as $u_{imp} = \sum_i{W_i u_i}$, where $u_i$ is the representation of keyword node which is obtained by the keyword graph learning procedure described above.
\textcolor{black}{In this way, with the help of neighbour nodes in the graph, we can obtain a more indistinguishable and difficult negative sample.}
The loss of keyword level contrastive learning thus can be written as:
\begin{equation}
    \mathcal{L}_{keyword} = -\E[log\frac{e^{h_(u_{in}, u_{out}) / \tau}}{\sum_{u_* \in U}e^{h(u_{in}, u_{*})/\tau}}],
\end{equation}
where $u_* \in U$ denotes the positive output keyword $u_{out}$ or imposter keyword $u_{imp}$. In keyword-level contrast, $h(\cdot)$ utilizes cosine similarity to calculate the distance between points.

\subsubsection{Mahalanobis Contrastive Learning}
\label{ma_dis}
Note that there exists a space gap between the instance-level contrast and the keyword-level contrast, which disturbs the completeness of this hierarchical contrastive architecture. 
Besides, the contrastive values vanish when the distance metric is hard to measure the actual discrepancy between positive and negative merely in instance distributions or in keyword representations.
To mitigate such problems, we design a Mahalanobis contrastive mechanism to correlate the instance distribution and keyword representation, where the objective is to minimize the margin between the output keyword $u_{out}$ and the posterior semantic distribution $q_\phi(z|x,y) \triangleq f_\phi(x,y)$ and maximize the margin between the imposter keyword $u_{imp}$ and the posterior distribution $f_\phi(x,y)$:
\begin{equation}
    \mathcal{L}_{ma} = -\E_{f_\phi}[log(1-\frac{e^{h(f_\phi(x,y), u_{out}) / \tau}}{\sum_{u_* \in U}e^{h(f_\phi(x,y), u_{*})/\tau}})],
\end{equation}
where $u_{*} \in U$ can be the positive output keyword $u_{out}$ or negative imposter keyword $u_{imp}$.
In Mahalanobis contrast, $h(\cdot)$ utilizes Mahalanobis distance~\citep{de2000mahalanobis} to measure the similarity from keyword point to the instance distribution.
In the univariate Gaussian case, $z \sim p(z|x,y)=N(\mu,\sigma^2)$, then the $h(f_\phi(x,y), u_*) \triangleq D_{MA}(p_\theta(z|x,y)|| u_{*}) = (u_{*} - \mu)\sigma^2I(u_{*} - \mu)$. 

Finally, we equip the CVAE model with the proposed hierarchical contrastive learning framework to unify hybrid granularities by adding $\mathcal{L}_{ins}$, $\mathcal{L}_{keyword}$ and $\mathcal{L}_{ma}$ to the reconstructed loss of Equation~\ref{cvae_rp}.
\section{Experiment}
\label{experiment}

\begin{table*}[t]
\centering
\small
\begin{tabular}{c|l|cccc|ccc}
\hline
&Models      & BLEU-1 & BLEU-2 & BLEU-3 & BLEU-4                 
& Extrema         & Average         & Greedy          \\ \hline
\multirow{9}{*}{(a).QQP}
& CVAE         & 0.4562          & 0.2150          & 0.0962          & 0.0496          & 0.6606          & 0.8371          & 0.8406          \\
& Seq2Seq     & 0.4510          & 0.2117          & 0.0950          & 0.0497          & 0.6543          & 0.8243          & 0.8533          \\
& Transformer & 0.4832          & 0.2339          & 0.1086          & 0.0590          & 0.6523          & 0.8274          & 0.8531          \\
\cline{2-9}
& Seq2Seq-DU  & 0.5613          & 0.2781          & 0.1334          & 0.0763          & 0.6679          & 0.8302          & 0.8590          \\
& DialoGPT   & 0.5749          & 0.2845          & 0.1337          & 0.0749          & 0.6658          & 0.8393          & 0.8597          \\
& BERT-GEN    & 0.5452          & 0.2781          & 0.1343          & 0.0764          & 0.6673          & 0.8299          & 0.8586          \\
& T5         & 0.6172 &	0.3301 &	0.1730 &	0.1019 &	0.6679 &	0.8408 &	0.8601          \\
\cline{2-9}
& Group-wise  & 0.5201          & 0.2472          & 0.1112          & 0.0582          & 0.6576          & 0.8337          & 0.8569          \\
& T5-CLAPS    & 0.6264 &	0.3394 &	0.1787 &	0.1058 &	0.6683 &	0.8430 &	0.8612          \\
\cline{2-9}
& Ours        & \textbf{0.6430} & \textbf{0.3517} & \textbf{0.1845} & \textbf{0.1153} & \textbf{0.6701} & \textbf{0.8495} & \textbf{0.8661} \\
\hline \hline

\multirow{9}{*}{(b).Douban}
& CVAE         & 0.0640 &	0.0259 &	0.0102 &	0.0047 &	0.4473 &	0.4814 &	0.6006      \\
& Seq2Seq     & 0.0542 &	0.0218 &	0.0086 &	0.0039 &	0.4388 &	0.4802 &	0.5960          \\
& Transformer & 0.0531 &	0.0210 &	0.0081 &	0.0036 &	0.4401 &	0.4807 &	0.5989        \\
\cline{2-9}
& Seq2Seq-DU  & 0.0887 &	0.0333 &	0.0123 &	0.0050 &	0.4591 &	0.4972 &	0.6083         \\
& Dialo-gpt   & 0.0953 &	0.0363 &	0.0136 &	0.0057 &	0.4633 &	0.4997 &	0.6108    \\
& BERT-GEN    & 0.0823  &	0.0314  &	0.0119  &	0.0050  &	0.4630  &	0.5001  &	0.6095          \\
& T5         & 0.1007 &	0.0379 &	0.0140 &	0.0057 &	0.4659 &	0.5056 &	0.6147        \\
\cline{2-9}
& Group-wise  & 0.0581 &	0.0239 &	0.0101 &	0.0053 &	0.4489 &	0.4903 &	0.6002  \\
& T5-CLAPS    & 0.1162 &	0.0413 &	0.0165 &	0.0062 &	0.4667 &	0.5071 &	0.6159 \\
\cline{2-9}
& Ours        & \textbf{0.1398} & \textbf{0.0516} & \textbf{0.0188} & \textbf{0.0074} & \textbf{0.4691} & \textbf{0.5091} & \textbf{0.6179} \\
\hline \hline

\multirow{9}{*}{(c).RocStories}
& CVAE         & 0.2581          & 0.0969         & 0.0360 & 0.0148          & 0.5135          & 0.5632          & 0.6011          \\
& Seq2Seq     & 0.2324          & 0.0896          & 0.0340 & 0.0150          & 0.5133        & 0.5649          & 0.6007          \\
& Transformer & 0.2552          & 0.0596          & 0.0354 & 0.0145          & 0.5129         & 0.5637          & 0.6006         \\
\cline{2-9}
& Seq2Seq-DU  & 0.3089          & 0.1131          & 0.0384 & 0.0156          & 0.5166         & 0.5773          & 0.6116          \\
& DialoGPT   & 0.3126          & 0.1105          & 0.0398 & 0.0158          & 0.5178      & 0.5860          & 6103          \\
& BERT-GEN    & 0.3040          & 0.1071          & 0.0385 & 0.0153          & 0.5173          & 0.5866          & 0.6098          \\
& T5         & 0.3274 &	0.1177 & 0.0416 & 0.0164          & 0.5194 &	0.5864 &	0.6118          \\
\cline{2-9}
& Group-wise  & 0.2748          & 0.1017          & 0.0366 & 0.0152          & 0.5139          & 0.5651          & 0.6009          \\
& T5-CLAPS    & 0.3420 &	0.1261 &	0.0452 &	0.0176 &	\textbf{0.5232} &	0.5880 &	\textbf{0.6134}        \\
\cline{2-9}
& Ours        & \textbf{0.3552} & \textbf{0.1341} & \textbf{0.0485} & \textbf{0.0184} & 0.5218 & \textbf{0.5884} & 0.6131 \\
\hline
\end{tabular}
\caption{Automatic evaluation results on (a) QQP for paraphrasing, (b) Douban for dialogue generation, and (c) RocStories for storytelling. The best results in each group are highlighted with \textbf{bold}.}
\label{tab:automatic_results}
\end{table*}

\subsection{Tasks and Datasets}
\label{task}
We conduct experiments on three public datasets QQP, Douban, RocStories for paraphrasing, dialogue generation, and storytelling task, respectively.
The details of the datasets are as follows:
\paragraph{Dialogue (Douban)}
Douban~\citep{group-wise} consists of Chinese daily conversations between pairs of speakers, collected from a popular social network website, Douban group\footnote{https://www.douban.com/}.
The dataset contains 218,039/10,000/10,000 context-response pairs for training/validation/test, with an average of 3.94 turns per context and 38.32 characters per utterance.
We concatenate historical dialogues and turn it into a single-turn dialogue training corpus.

\paragraph{Paraphrasing (QQP)}
QQP~\citep{WinNT, wang2019glue} is a dataset published by the community question-answering website Quora on whether a pair of questions is semantically consistent.
To adapt it to the contrastive learning task, we only keep question pairs that have positive and negative rewriting for the same input.
Thus, there remain 44,949 samples in the dataset, which are split into training/validation/test sets of 40,441/2,254/2,254 samples.

\paragraph{Storytelling (RocStories)}
RocStories consists of 98,163 high-quality hand-crafted stories, which capture causal and temporal commonsense relations of daily events~\citep{mostafazadeh2016corpus}.
Each story paragraph contains 5 sentences with an average of 43 words.
Following the previous work \citet{yu2021content}, we split the dataset into 8:1:1 for training, validation, and test.

For the above three datasets, in order to construct different levels of contrastive learning, we performed the same preprocessing of extracting keywords.
We utilize the TextRank model~\citep{mihalcea2004textrank} to extract keywords from each input and output sample, respectively.
Besides, the vocabulary size of both datasets is the same as BERT \citep{bert} setting.

\subsection{Implementation Details}
Our experiments are implemented in Tensorflow~\citep{Abadi2016TensorFlowAS} on an NVIDIA Tesla P100 GPU.
For our model and all baselines, we follow the same setting as described below.
We pad or cut the input to 100, 20, 100 words for dialogue generation, paraphrasing, and storytelling, respectively.
The truncation length is decided based on the observation that there is no significant improvement when increasing input length.
The minimum decoding step is 5, and the maximum step is 20 for all tasks.
Experiments were performed with a batch size of 256, and
we use Adam optimizer~\citep{kingma2014adam} as our optimizing algorithm.
During the test stage, the beam-search size is set to 4 for all methods and the checkpoint with the smallest validation loss is chosen.
Note that for better performance, our model is built based on BERT, and the decoding process is the same as Transformer~\citep{transformer}.
Finally, due to the limitation of time and memory, small settings are used in the pre-training baselines.

\begin{table*}[ht]
    \centering
    \small
    \begin{tabular}{l|cccc|ccc}
    \hline
    Models      & BLEU-1 & BLEU-2 & BLEU-3 & BLEU-4                                       & Extrema         & Average         & Greedy          \\ \hline
    ours        & \textbf{0.6430} & \textbf{0.3517} & \textbf{0.1845} & \textbf{0.1153} & \textbf{0.6701} & \textbf{0.8495} & \textbf{0.8661} \\
    w/o graph       &  0.6295 &	0.3333	& 0.1675	& 0.1001	& 0.6685	& 0.8455	& 0.8647\\
    w/o keyword     &   0.5764	&	0.2993	&	0.1499	&	0.0892	&	0.6673	&	0.8450	&	0.8539\\
    w/o MA      &   0.6013	& 0.3208	& 0.1628	& 0.0981	& 0.6605	& 0.8436	& 0.8524 \\
    \hline
    \end{tabular}
    \caption{Ablation results on dataset QQP.}
    \label{tab:ablation_qqp}
\end{table*}

\subsection{Compared Baselines}
We compare our method against several traditional generation models, pretrained-based generation models, and contrastive learning models.

\emph{Traditional generation models:}
(1) \textbf{CVAE}~\citep{zhao2017learning} generates sentences based on latent variables, sampling from potential semantic distribution.
(2) \textbf{Seq2Seq}~\citep{seq2seq} is a sequence-to-sequence framework combined with attention mechanism and pointer network.
(3) \textbf{Transformer}~\citep{transformer} is an abstractive method based solely on attention mechanisms.

\emph{Pretrained-based generation models:}
(4) \textbf{Seq2Seq-DU}~\citep{feng2020sequence} is concerned with dialogue state tracking in a task-oriented dialogue system.
(5) \textbf{DialoGPT}~\citep{dialogpt} proposes a large, tunable neural conversational response generation model trained on more conversation-like exchanges.
(6) \textbf{BERT-GEN}~\citep{bert} augments Seq2Seq with BERT as the encoder.
(7) \textbf{T5}~\citep{t5} introduces a unified framework that converts all text-based language problems into a text-to-text format.

\emph{Contrastive learning methods:}
(8) \textbf{Group-wise}~\citep{group-wise} augments contrastive dialogue learning with group-wise dual sampling.
(9) \textbf{T5-CLAPS}~\citep{Claps} generates negative and positive samples for contrastive learning by adding small and large perturbations, respectively.

\subsection{Evaluation Metrics}
To evaluate the performance of our model against baselines, we adopt the following metrics widely used in existing studies.
\paragraph{BLEU}
We utilize BLEU score \citep{papineni2002bleu} to measure word overlap between the generated text and the ground-truth.
Specifically, following the conventional setting of \cite{gu2018dialogwae}, we adopt BLEU-1$\sim$4 scores under the smoothing techniques (smoothing 7).

\paragraph{Embedding}
To evaluate our model more comprehensively, we also capture the semantic matching degrees between the bag-of-words (BOW) embeddings of generated text and reference~\citep{gu2018dialogwae}.
Particularly we adopt three metrics:
1) \textit{Extrema}, cosine similarity between the largest extreme values among the word embeddings in the two texts;
2) \textit{Average}, cosine similarity between the averaged word embeddings of generated text and reference;
3) \textit{Greedy}, greedily matching words in the two texts based on cosine similarities.

\subsection{Experimental Results}

\begin{table}[h]
\small
\centering
    \begin{tabular}{lcccc}
      \hline
      & Flu & Mean & Diff \\
      \hline
      Seq2Seq-DU & 2.03 & 2.12  & 1.76   \\
      DialoGPT & 2.18  &  2.04 & 1.97 \\
      T5-CLAPS & 2.24  &  2.16 & 2.19 \\
      Ours & \textbf{2.51}\dubbelop &\textbf{2.45}\dubbelop  & \textbf{2.43}\dubbelop \\
      \hline
    \end{tabular}
    \caption{Fluency (Flu), Meaningfulness (Mean), and Differential (Diff) comparison by human evaluation. \dubbelop shows the statistical significance tested by a two tailed paired t-test.}
    \label{tab:human}
\end{table}

\subsubsection{Overall Performance}
\paragraph{Automatic Evaluation}
The experimental results ars summarized in  Table~\ref{tab:automatic_results}.
The upper part lists the effects of traditional generation methods such as \texttt{Seq2Seq} and \texttt{Transformer}, and the lower part shows the latest pretrained-based methods including \texttt{DialoGPT} and \texttt{T5}.
Overall, pretrained-based methods generally outperform traditional methods, and this also proves the effectiveness of the pretrained language model on the generation tasks.
Secondly, we can find that the performance is significantly improved after adding contrast learning.
Finally, our method outperforms \texttt{T5-CLAPS} by 2.7\%, 3.6\% on QQP, by 20.3\%, 24.9\% on Douban, and by 3.9\%, 6.3\% on RocStories in terms of BLEU-1, BLEU-2, respectively, which proves the superiority of our model.

\paragraph{Human Evaluation}
We also assessed system performance by eliciting human judgments on 100 randomly selected test instances on QQP dataset.
Three annotators are asked to rate paraphrasing questions generated by \texttt{T5-CLAPS}, \texttt{DialoGPT}, \texttt{Seq2Seq-DU}, and our model according to Fluency (Flu), Meaningfulness (Mean), and Differential (Diff).
The rating score ranges from 1 to 3, with 3 being the best.
Table~\ref{tab:human} lists the average scores of each model, showing that our model outperforms other baselines among all metrics, which indicates that our model generates paraphrasing sentences more readable successfully.
The kappa statistics are 0.53, 0.61, and 0.56 for fluency, meaningfulness, and differential, respectively, which indicates the moderate agreement between annotators.

\begin{table}[t]
\centering
\small
\begin{tabular}{c|l}
  \toprule
  \multicolumn{2}{p{7.2cm}}{    \emph{\textbf{Input text:}}
    What one exercise will help me lose belly fat?
  } \\ \hline 
  
  \multicolumn{2}{p{7.2cm}}{
    \emph{\textbf{Reference paraphrasing text:}}
    How do i remove belly fat?
  }\\ \hline
  \multicolumn{2}{p{7.2cm}}{
    \textbf{Keyword extracted by TextRank:  belly fat}
  }\\
  \multicolumn{2}{p{7.2cm}}{
    \emph{\textbf{Generated text 1:}}
    What is the best exercise way to lose \hlc[pink]{belly fat}?
  }\\ \hline
   \multicolumn{2}{p{7.2cm}}{
    \textbf{Keyword from random-selected: disposable}
  }\\
  \multicolumn{2}{p{7.2cm}}{
    \emph{\textbf{Generated text 2:}}
    Can \hlc[cyan!50]{one-off exercise} lose belly fat?
  }\\ \hline

\multicolumn{2}{p{7.2cm}}{
    \emph{\textbf{Seq2Seq-DU:}}
    What are the best ways to lose weight?
    }\\ \hline
\multicolumn{2}{p{7.2cm}}{
    \emph{\textbf{DialoGPT:}}
    Which exercise helps to lose weight?
    }\\ \hline
  \multicolumn{2}{p{7.2cm}}{
    \emph{\textbf{T5-CLAPS:}}
    How can I lose weight?
    }\\
  \bottomrule
\end{tabular}
\caption{Case study to verify the influence of sampling different keywords.
The texts in red and blue indicate the parts corresponding to the extracted keyword and random-selected keyword, respectively.
}
\label{tab:sample}
\end{table}

\subsubsection{Ablation Study}
We conduct ablation tests to assess the importance of the keyword graph architecture (\texttt{w/o graph}), keyword (\texttt{w/o keyword}), as well as the Mahalanobis contrast (\texttt{w/o MA contrast}), and the results are shown in  Table~\ref{tab:ablation_qqp}.
Concretely, after removing the keywords (w/o keyword), using only instance-level contrastive, the effect of our model is greatly reduced by about 10.4\%, which illustrates the desirability of considering the contributions of words in a sentence.
On this basis, adding keyword contrastive learning with removing the keyword graph, the effect of the model has been improved but is still lower than our model by 2.1\%.
This shows that keywords are indeed conducive to capturing important information, and it also illustrates the significance of a keyword graph.
Finally, the experiment of removing the Mahalanobis contrastive loss indicates that only with granularity independent contrast is not sufficient, and the Mahalanobis contrast plays a critical intermediate role.

\subsubsection{Visualization of Different Levels of Contrastive Learning}
\begin{figure}[h]
  \centerline{\includegraphics[width=6.8cm]{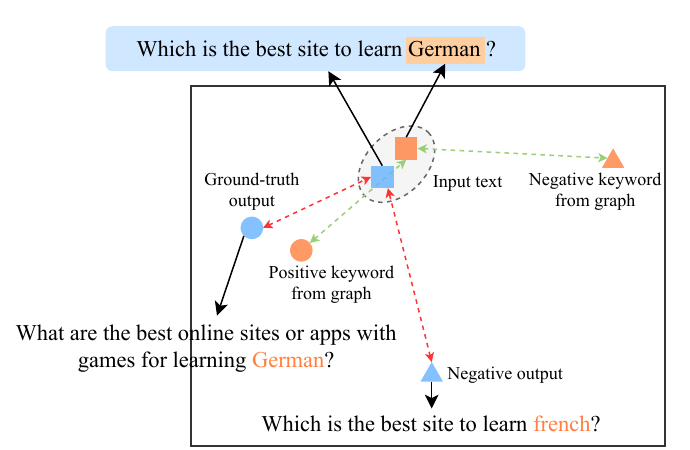}}
  \caption{\textcolor{black}{Visualization of contrastive learning.
The square, circle and triangle represents the input text, positive output sample, and negative output sample, respectively.
Blue represents the sentence, and yellow represents the keyword.
}}
  \label{fig:visual}
\end{figure}

To study the hierarchical contrastive learning, we visualize the vectors of keyword, input text, positive and negative output text on randomly sampled cases from QQP dataset, as shown in Figure~\ref{fig:visual}.
For visualization purposes, we reduce the dimension of the latent vector with t-SNE \citep{maaten2008visualizing}.
It can be observed that the input sentence representation is located close to the keyword, which shows that the keyword, as the most important information in the sentence, determines the semantic distribution.
Moreover, in contrastive learning, it can be seen that after training, the position of the input sentence is close to the positive samples and far away from the negative samples.
This suggests that contrastive learning can correct the semantic distribution.

\subsubsection{Analysis of Different Keywords}
We finally investigate the influence of sampling different keywords.
As shown in Table~\ref{tab:sample}, for an input question, we provide keywords extracted by TextRank and randomly-selected keywords as the condition to control the semantic distribution and examine the quality of the generated text.
As the most important information unit, different keywords lead to different semantic distributions and will result in different generated texts.
The more properly the keywords are selected, the more accurately the sentences will be generated.
When utilizing the keywords extracted by TextRank as a condition, the information ``belly fat'' is focused during the generation of paraphrasing questions, and the generated sentences are more accurate.
On the contrary, after adding the random-selected keyword ``disposable'', the generated question emphasizes ``one-off exercise'', which brings incorrect information.

We also compare our model with several baselines in Table~\ref{tab:sample}.
Most baselines can generate fluent questions in this case.
However, they focus on ``lose weight'', and miss the significant information ``belly fat''.
Based on the above analysis, we can observe that keywords can emphasize and protect the highlight information in sentences, and affect the semantic distribution of as a condition.

\section{Conclusion}
In this paper, we propose a hierarchical contrastive learning mechanism, which consists of intra-contrasts within instance-level and keyword-level and inter-contrast with Mahalanobis contrast.
The experimental results yield significant out-performance over baselines when applied in the CVAE framework. 
In the future, we aim to extend the contrastive learning mechanism to different basic models, and will explore contrastive learning methods based on external knowledge.

\section{Acknowledgments}
We would like to thank the anonymous reviewers for their constructive comments. 
This work was supported by National Key Research and Development Program of China (No. 2020AAA0106600), National Natural Science Foundation of China (NSFC Grant No. 62122089, No. 61832017 \& No. 61876196), and Beijing Outstanding Young Scientist Program No. BJJWZYJH012019100020098. This work was also supported by Alibaba Group through Alibaba Research Intern Program.

\section{Ethics Impact}
In this paper, we propose an inter-level contrastive learning method, which unifies instance-level and keyword-level contrasts in the CVAE framework. The positive impact lies in that it can help improve the capability of generation models on paraphrasing, dialogue generation, and storytelling tasks. The negative impact may be that the generation process of the system is not fully controllable, so it is possible to generate inaccurate or unreasonable content in some extreme cases. Hence, extra processing steps might be needed if this method were to be used in scenarios where high accuracy is required.



\end{document}